\documentclass[twoside,twocolumn,10pt]{article}
\usepackage{fancyhdr}
\usepackage{geometry}
\usepackage{multicol}
\usepackage{abstract}
\usepackage{graphicx}
\usepackage{titlesec}
\usepackage{ragged2e}
\usepackage{amssymb}
\usepackage{parskip}
\usepackage{amsmath}
\usepackage{newtxtext}
\usepackage{booktabs}
\usepackage{array}
\usepackage{tabularx}
\usepackage{url}
\usepackage[hidelinks]{hyperref}
\usepackage{microtype}

\newcolumntype{L}[1]{>{\raggedright\arraybackslash}p{#1}}
\newcolumntype{C}[1]{>{\centering\arraybackslash}p{#1}}
\newcolumntype{R}[1]{>{\raggedleft\arraybackslash}p{#1}}

\usepackage[font=footnotesize, labelfont=bf, justification=raggedright, format=plain]{caption}

\usepackage[backend=bibtex,style=ieee]{biblatex}
\defbibheading{bibliography}[\refname]{%
  \section{\MakeUppercase{REFERENCES}}}

\titleformat{\section}{\large\bfseries\uppercase}{\thesection.}{1em}{}
\titleformat{\subsection}{\normalfont\bfseries\flushleft}{\thesubsection.}{1em}{}
\titleformat{\subsubsection}{\normalfont\bfseries\itshape\flushleft}{\thesubsubsection}{1em}{}

\newcommand{\keywordsname}{Keywords}

\fancypagestyle{firstpage}{
  \fancyhf{}
  \fancyhead[LO]{\small Volume X, Issue Y, Year}
  \fancyhead[RO]{\small J. Comput. Sci. Appl.}
  \fancyfoot[L]{\footnotesize Received: Month Year; Revised: Month Year; Accepted: Month Year\\\textsuperscript{*}Corresponding Author: Wuming Lei \textless WMFCDS@outlook.com\textgreater}
  \fancyfoot[R]{\footnotesize \copyright\ Year by the Author(s). Published by JCSA,\\\textsuperscript{*} under the CC-BY license}

}

\title{\fontsize{16}{20}\selectfont\textbf{A QUBO-Inspired Computational Framework for Airport Landside Bottleneck Diagnosis and Dynamic Dispatch Optimization}}
\author{
    \fontsize{12}{14}\selectfont
    Wuming Lei\textsuperscript{1*}, Xiaobin Li\textsuperscript{1}, Mingyan Sun\textsuperscript{1}, Jianing Long\textsuperscript{1}, Yulin Tong\textsuperscript{1}, Yanbin Gao\textsuperscript{1}\\[1ex]
    \fontsize{12}{14}\selectfont
    \textsuperscript{1}East China Jiaotong University, Nanchang, China
}
\date{}

\begin{document}

\twocolumn[
\begin{@twocolumnfalse}
\vspace{-0.5cm}
\raggedright
{\fontsize{11}{13}\selectfont{Research Article}}
\vspace{-0.8cm}
\maketitle
\thispagestyle{firstpage}

\begin{abstract}
\fontsize{11}{13}\selectfont
\vspace{1em}
\justifying
Airport landside traffic centers connect terminal arrivals with taxis, ride-hailing vehicles, private cars, buses, metro services, parking facilities, and terminal-area roadways. Peak arrivals can create coupled congestion across passenger queues, vehicle queues, pickup berths, storage areas, and access roads. This study proposes a QUBO-inspired computational framework for bottleneck diagnosis and dynamic dispatch in this setting. Shanghai Pudong International Airport and Hangzhou Xiaoshan International Airport serve as case airports. A five-minute state model links passenger arrivals, vehicle supply, pickup berth service, vehicle storage, and road capacity. Bottleneck diagnosis uses service intensity, road demand saturation, bottleneck frequency, queue severity, shadow-price leverage, and a composite congestion severity index. Two dispatch schemes are tested under consistent demand inputs: finite-action model predictive control and quadratic-unconstrained-binary-optimization-inspired simulated annealing. In the strong-peak baseline scenario, the QUBO-inspired method reduces the final passenger queue from 3445 to 2477 passengers at Shanghai Pudong and from 2053 to 1482 passengers at Hangzhou Xiaoshan. Case results indicate different dominant bottlenecks. Shanghai Pudong is more affected by road saturation, whereas Hangzhou Xiaoshan is more affected by pickup berth service. Robustness tests under demand, supply, service, road-capacity, modal-share, and random-noise perturbations show retained queue-reduction benefits under the tested uncertainty levels.

\vspace{0.3em}
\noindent Source code, processed inputs, experiment outputs, and figure-generation scripts are available at \url{https://github.com/Ming23233/airport-landside-qubo-dispatch}.

\vspace{-0.1cm}
\noindent\textbf{keywords}
Airport landside traffic; bottleneck diagnosis; passenger queue; model predictive control; QUBO-inspired simulated annealing; robustness analysis.
\end{abstract}

\vspace{1cm}
\end{@twocolumnfalse}
]

\section{Introduction}
\fontsize{12}{14}\selectfont
\justifying
Peak-period airport landside operations involve several tightly coupled service processes. Passenger arrivals, taxis, ride-hailing vehicles, private cars, buses, parking facilities, pickup berths, and terminal-area roads interact over short time intervals. Recent airport guidance treats curbside and terminal-area roadway operations as a distinct planning and control problem rather than a minor terminal-access detail \cite{ACRP266}. Emerging mobility services have further changed airport access patterns and increased the need for adaptive curbside management \cite{ACRP286}.

Congestion at an airport landside traffic center may arise from different binding resources. A terminal queue can be caused by insufficient vehicle supply, inadequate pickup berth service, limited storage space, or downstream road saturation. Operational conclusions can therefore change if these constraints are analyzed separately. The proposed framework identifies the active bottleneck and evaluates dispatch policies under the same state representation.

An integrated framework is developed for this purpose. Demand construction, state propagation, bottleneck diagnosis, dispatch optimization, and robustness testing are kept in one workflow. Shanghai Pudong International Airport and Hangzhou Xiaoshan International Airport are used as case airports. Their traffic scales and facility characteristics differ, which supports a comparative analysis of road-dominated and berth-dominated congestion mechanisms.

The study makes three contributions. First, a passenger-vehicle-facility-road state model is formulated at a five-minute time step. Passenger queues, vehicle queues, pickup berth service, storage capacity, and road capacity are updated in the same simulation loop. Second, a diagnostic layer distinguishes vehicle-supply, berth-facility, demand-limited, and road-saturation bottlenecks. Third, finite-action model predictive control (MPC) and a quadratic-unconstrained-binary-optimization (QUBO) inspired simulated annealing method are compared under consistent inputs. Interpretable queue, waiting, saturation, and robustness indicators are reported.

\begin{figure*}[t]
\centering
\includegraphics[width=\textwidth]{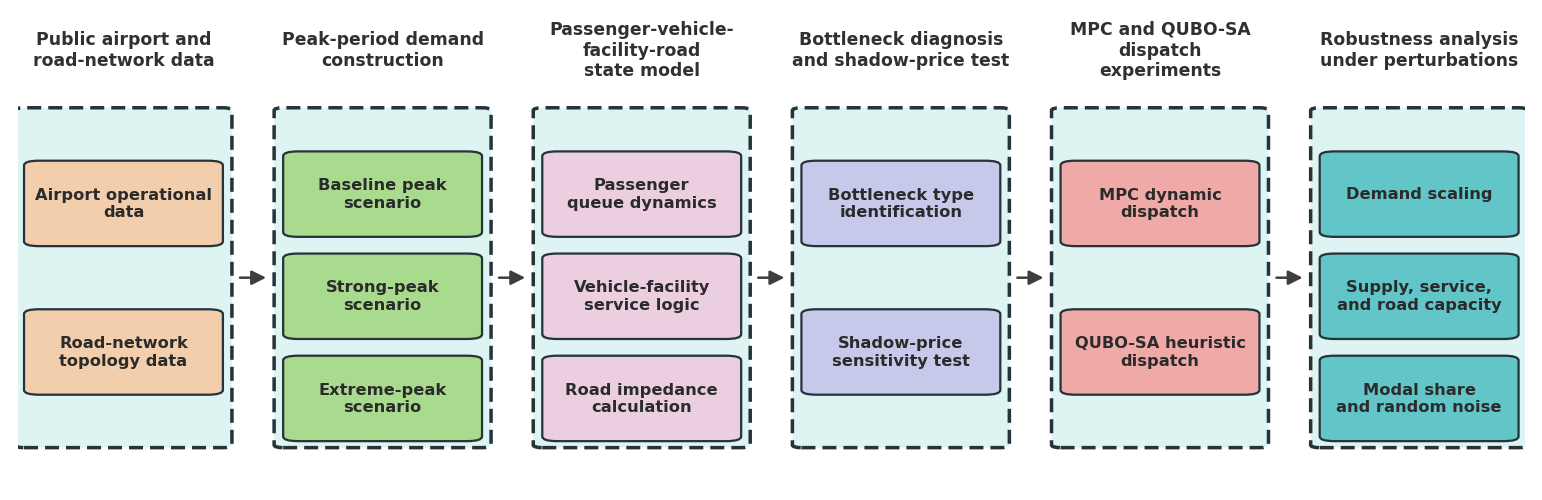}
\caption{Detailed computational workflow for airport landside bottleneck diagnosis, dispatch evaluation, and robustness testing.}
\label{fig:framework}
\end{figure*}

Figure~\ref{fig:framework} summarizes the computational workflow used in the case-study experiments.

\section{Literature Review}
Airport curbside management has been examined through facility design, traffic simulation, vehicle routing, demand management, and curb allocation. A recent airport-focused modeling framework evaluates curbside traffic-management policies at Dallas-Fort Worth International Airport \cite{Ugirumurera2021}. Mesoscopic airport curbside simulation also provides a practical way to represent passenger and vehicle interactions without relying on a fully microscopic model \cite{Harris2017}.

Airport access has been reshaped by ride-hailing, transit availability, and emerging mobility. Evidence from New York shows that ride-hailing can alter taxi and transit shares in airport ground access \cite{Dong2021}. Transit payment data have also been used to infer airport access and airport-choice behavior \cite{Wang2022TransportPolicy}. Passenger-level access-mode studies in European airports further show that access decisions vary with traveler attributes and trip context \cite{Colovic2022}. Sustainability-oriented reviews highlight the importance of landside access in airport decarbonization strategies \cite{Mahesh2025}.

Curbside control studies increasingly consider pricing, real-time allocation, and short-horizon management. Optimal pricing has been formulated for managing ride-hailing pickup and drop-off activity at the curb \cite{Liu2023CurbPricing}. Surrogate-based real-time curbside management has also been proposed for ride-hailing and delivery operations \cite{Vishnoi2025}. These studies motivate a dispatch-oriented view of airport landside operations. This study combines bottleneck diagnosis with dispatch comparison and robustness testing in one case-study framework.

\section{Data and Scenario Design}
\subsection{Case Airports and Data Sources}
Shanghai Pudong International Airport and Hangzhou Xiaoshan International Airport are selected as case airports. Shanghai Pudong represents a large hub with high passenger throughput and substantial road pressure. Hangzhou Xiaoshan represents a large regional hub where pickup service capacity can become influential during strong arrival peaks. Public airport throughput statistics provide daily passenger and aircraft movement indicators \cite{CAAC2025}. Geofabrik OpenStreetMap extracts provide the surrounding road-network topology \cite{Geofabrik2026}.

Input variables are organized into three groups. Airport-scale indicators include daily passengers, daily movements, and passengers per movement. Facility variables include taxi berths, ride-hailing pickup areas, parking capacity, and vehicle storage assumptions. Road-network variables include nodes, edges, average degree, and road class composition. Table~\ref{tab:airport-inputs} summarizes the main inputs used in the experiments.

\begin{table*}[t]
\centering
\caption{Case-airport input summary used in the computational experiments.}
\label{tab:airport-inputs}
\resizebox{\textwidth}{!}{%
\begin{tabular}{C{2.6cm}C{2.0cm}C{1.9cm}C{2.2cm}C{1.7cm}C{2.2cm}C{1.9cm}C{1.6cm}C{1.6cm}}
\toprule
Airport & Daily passengers & Daily movements & Passengers per movement & Taxi berths & Ride-hailing zones/points & Parking capacity & Road nodes & Road edges \\
\midrule
Shanghai Pudong & 232861.8 & 1526.1 & 152.58 & 20 assumed & 90 points & 10023 & 1865 & 1765 \\
Hangzhou Xiaoshan & 138243.9 & 902.3 & 153.21 & 18 observed & 3 zones & 4400 & 2474 & 2158 \\
\bottomrule
\end{tabular}%
}
\end{table*}

\subsection{Peak Demand Construction}
\begin{figure}[h]
\centering
\includegraphics[width=\columnwidth]{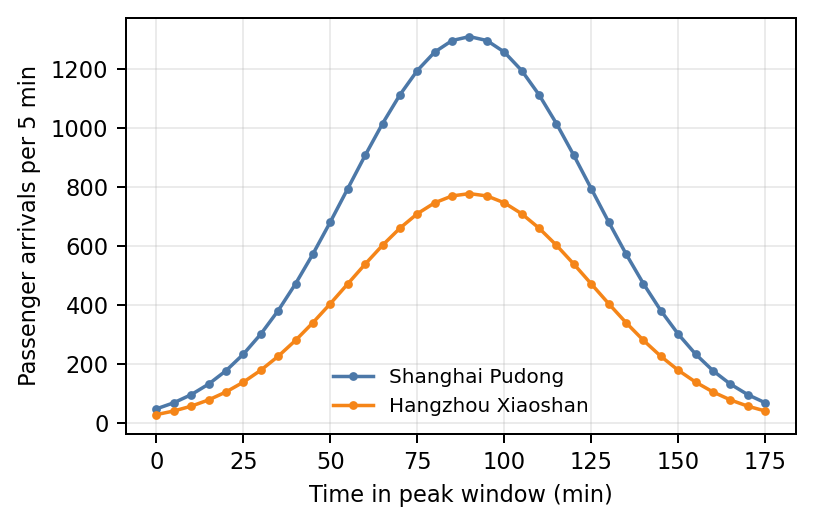}
\caption{Constructed passenger arrival profiles in the strong-peak baseline scenario.}
\label{fig:demand-curve}
\end{figure}

All experiments use a three-hour peak window with a five-minute step. Passenger arrivals follow a normalized Gaussian profile. Peak-hour demand in the constructed profile is calibrated to the selected peak-hour share of daily arriving passengers. Three peak scenarios are considered: normal peak, strong peak, and extreme peak. Three modal-share scenarios are also considered: baseline, car-intensive, and public-transport-intensive. Figure~\ref{fig:demand-curve} shows the constructed arrival profile for the strong-peak baseline scenario.

\section{Methodology}
\subsection{Integrated State Model}
Let $t$ denote a five-minute time step and let $m$ denote a landside transport mode. Modeled vehicle modes include taxi, ride-hailing vehicle, private car, and bus. Metro and other modes are included in the passenger demand split, but they do not create vehicle pickup queues in the landside traffic center model.

For each vehicle mode, the state includes the passenger queue $Q^p_{m,t}$, the vehicle queue $Q^v_{m,t}$, and the road queue $Q^r_{m,t}$. Step demand includes passenger arrival $a^p_{m,t}$ and vehicle arrival $a^v_{m,t}$. Vehicle occupancy is denoted by $\eta_m$. Terminal-side service before roadway merging is constrained by vehicle availability, facility service capacity, and passenger availability:
\begin{equation}
s^{0}_{m,t}=\min \left(Q^v_{m,t}+a^v_{m,t}, C^b_{m,t}, \frac{Q^p_{m,t}+a^p_{m,t}}{\eta_m} \right),
\end{equation}
where $C^b_{m,t}$ is the effective berth or curb service capacity. Taxi, ride-hailing vehicle, and bus use pickup berths. Private cars use a curb-flow representation.

Roadway discharge is represented as a shared downstream constraint. Let $C^r_t$ be road capacity per step. Loaded vehicles from all modes compete for this capacity. Carried-over road queues are released first. New loaded vehicles then share the remaining capacity proportionally. This mechanism keeps realized road saturation at or below capacity while still recording the demand-side pressure:
\begin{equation}
\sigma^d_t=\frac{\sum_m (Q^r_{m,t}+s^{0}_{m,t})}{C^r_t}.
\end{equation}
Realized saturation and demand saturation have different diagnostic roles. Demand saturation may exceed one, and it is used to identify over-saturation pressure.

Passenger waiting and vehicle waiting are estimated from queue-service ratios. Little's law provides the queueing consistency basis for relating average queue length, arrival rate, and waiting time \cite{Little1961}. The roadway impedance layer uses the Bureau of Public Roads volume-delay form as an engineering approximation for the relation between volume-capacity ratio and speed degradation \cite{BPR1964}. The dynamic road-capacity representation is compatible with the traffic-flow idea that downstream supply can constrain upstream discharge, which is also central to cell-transmission modeling \cite{Daganzo1994}.

\subsection{Bottleneck Diagnosis}
Four bottleneck labels are used:
\begin{itemize}
    \item vehicle supply, when available vehicles are the binding constraint;
    \item berth facility, when pickup berth or curb service is the binding constraint;
    \item demand limited, when passenger availability is lower than service and vehicle capacity;
    \item road saturation, when total loaded vehicle demand exceeds the downstream road capacity.
\end{itemize}

For each resource $r$, the bottleneck frequency $F_r$ is the share of time-mode cells in which $r$ is the active bottleneck. Severity $S_r$ is the accumulated positive passenger-queue increment associated with that bottleneck. A shadow-price test perturbs each resource by 10\% and measures the reduction in total weighted delay. The composite congestion severity index (CSI) combines mean road over-saturation intensity, peak passenger backlog relative to a design queue capacity, and average passenger waiting relative to a service-promise threshold.

\subsection{Dynamic Dispatch Schemes}
Finite-action MPC is used as the first dispatch scheme. At each time step, the controller evaluates a set of candidate actions over a six-step horizon. Each action specifies pickup berth allocation, release factors, and modal guidance shifts. The cost includes passenger queues, vehicle queues, passenger waiting, driver waiting, road congestion, fairness, guidance penalty, storage overflow, and rewards for served passengers and vehicles. MPC is suitable for constrained dynamic control because it handles receding horizons and explicit operational constraints \cite{Mayne2000}. Traffic-control applications of MPC provide a related methodological basis for short-horizon roadway operations \cite{Ye2019}.

QUBO-inspired simulated annealing is used as the second dispatch scheme. The dispatch decision is discretized into berth allocation choices, release-factor choices, and modal-shift choices. The energy function has the same operational components as the MPC cost. QUBO models provide a compact way to represent discrete optimization problems in binary form \cite{Glover2022}. Binary formulations of combinatorial problems are closely related to Ising-model representations \cite{Lucas2014}. Simulated annealing is then used as the search procedure for the QUBO-inspired energy landscape \cite{Kirkpatrick1983}. The method is quantum-inspired and does not require quantum hardware.

\section{Experimental Design}
Three experiment groups are conducted. Bottleneck diagnosis is first performed under the strong-peak baseline scenario. Dispatch performance is then compared for no dispatch, MPC, and QUBO-inspired simulated annealing under all peak and modal-share scenarios. Robustness is finally tested by perturbing demand, service time, road capacity, vehicle supply, modal share, and random demand noise.

Table~\ref{tab:scenario-design} lists the main scenario settings. The robustness experiment keeps the dispatch logic fixed and perturbs the environment. This design is intended to test whether a dispatch scheme remains useful when the realized operating condition differs from the nominal case.

\begin{table}[h]
\centering
\caption{Scenario settings used in the numerical experiments.}
\label{tab:scenario-design}
\resizebox{\columnwidth}{!}{%
\begin{tabular}{C{2.9cm}C{5.3cm}}
\toprule
Component & Setting \\
\midrule
Time step & 5 min \\
Simulation horizon & 3 h \\
Peak scenarios & Normal, strong, extreme \\
Modal-share scenarios & Baseline, car-intensive, public-transport-intensive \\
Dispatch methods & No dispatch, MPC, QUBO-inspired SA \\
MPC horizon & 6 steps \\
Robustness factors & Demand, supply, service time, road capacity, modal share, noise \\
\bottomrule
\end{tabular}%
}
\end{table}

\section{Results and Discussion}
\subsection{Bottleneck Diagnosis}
Bottleneck diagnosis indicates different dominant mechanisms at the two airports. At Shanghai Pudong, road saturation is the main source of modeled pressure. Road demand saturation reaches 3.892 in the strong-peak baseline diagnostic model, and road saturation accounts for 63.9\% of the time-mode bottleneck cells. The modeled speed index falls to 1.13 km/h under the peak pressure state. This value is interpreted as an internal pressure indicator derived from the BPR function, not as an observed field speed.

\begin{figure}[h]
\centering
\includegraphics[width=\columnwidth]{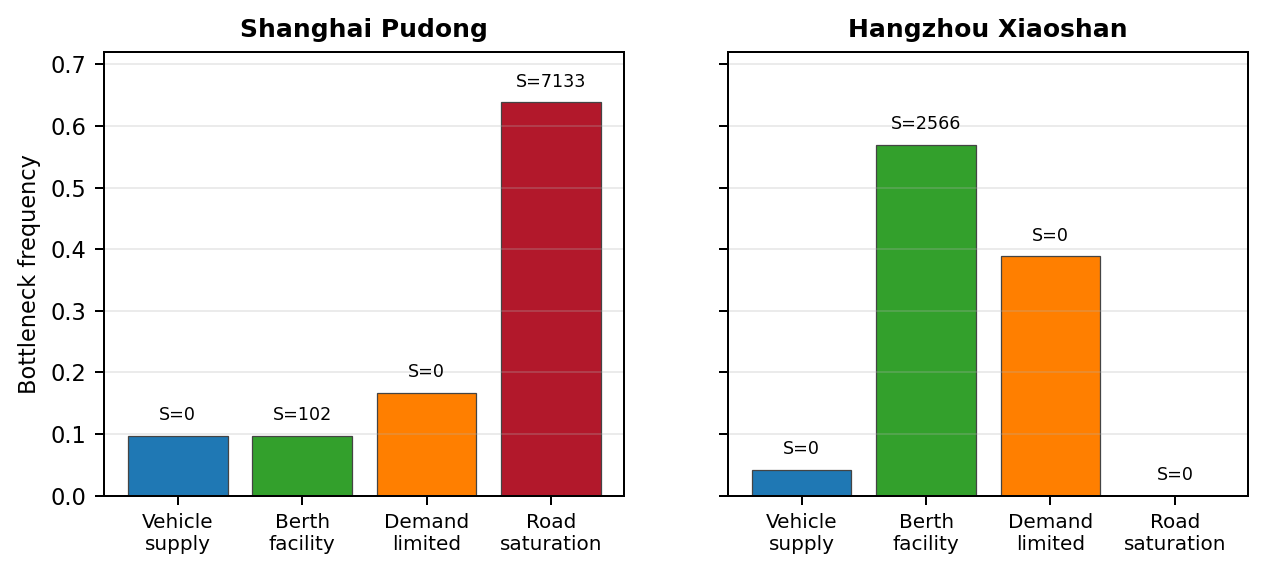}
\caption{Bottleneck frequency and queue-severity annotations in the strong-peak baseline scenario.}
\label{fig:bottleneck}
\end{figure}

At Hangzhou Xiaoshan, berth facility pressure is more prominent. The ride-hailing service intensity reaches 6.049, and the berth-facility bottleneck accounts for 56.9\% of time-mode cells. The road demand saturation remains below one in the strong-peak baseline diagnostic experiment. This suggests that improving pickup service capacity and vehicle-passenger matching may provide a more direct intervention path than roadway expansion for this case setting. Figure~\ref{fig:bottleneck} visualizes the bottleneck frequencies and severity annotations, and Table~\ref{tab:bottleneck-summary} reports the diagnostic summary.

\begin{table}[h]
\centering
\caption{Bottleneck diagnosis summary in the strong-peak baseline scenario.}
\label{tab:bottleneck-summary}
\resizebox{\columnwidth}{!}{%
\begin{tabular}{C{2.8cm}C{1.4cm}C{2.2cm}C{3.6cm}}
\toprule
Airport & CSI & Migration count & Main diagnostic result \\
\midrule
Shanghai Pudong & 10.2496 & 8 & Road saturation pressure \\
Hangzhou Xiaoshan & 4.0671 & 4 & Berth facility pressure \\
\bottomrule
\end{tabular}%
}
\end{table}

\subsection{Dispatch Comparison}
Under the strong-peak baseline scenario, QUBO-inspired simulated annealing reduces the final passenger queue from 3445.4 to 2476.5 passengers at Shanghai Pudong. Average passenger waiting decreases from 39.74 min to 25.71 min. Average driver waiting decreases from 36.73 min to 31.14 min. MPC also reduces the final passenger queue at Shanghai Pudong, but its reduction is smaller in this experiment.

\begin{figure}[h]
\centering
\includegraphics[width=\columnwidth]{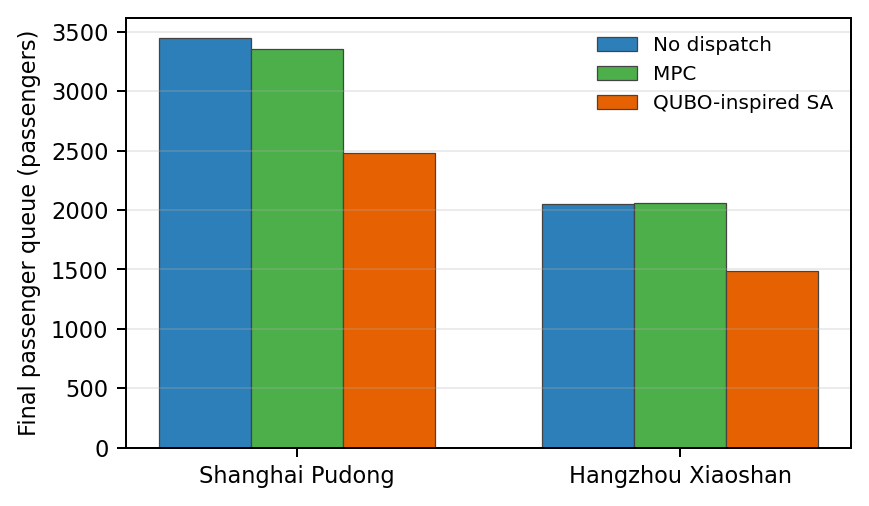}
\caption{Final passenger queue comparison under the strong-peak baseline scenario.}
\label{fig:dispatch}
\end{figure}

At Hangzhou Xiaoshan, QUBO-inspired simulated annealing reduces the final passenger queue from 2053.3 to 1481.9 passengers. The average passenger waiting estimate decreases from 51.71 min to 28.26 min. The average driver waiting estimate decreases from 54.85 min to 36.53 min. MPC reduces waiting time at Hangzhou Xiaoshan, but the final passenger queue is close to the no-dispatch value. This outcome is consistent with a berth-facility bottleneck. A small shift in berth allocation and release factors can reduce waiting, but full queue clearance remains limited by service capacity. Figure~\ref{fig:dispatch} and Table~\ref{tab:dispatch-performance} summarize the dispatch comparison.

\begin{table*}[t]
\centering
\caption{Dispatch performance under the strong-peak baseline scenario.}
\label{tab:dispatch-performance}
\resizebox{\textwidth}{!}{%
\begin{tabular}{C{2.7cm}C{2.5cm}C{2.4cm}C{2.7cm}C{2.7cm}C{3.1cm}}
\toprule
Airport & Method & Final passenger queue & Average passenger wait (min) & Average driver wait (min) & Maximum road demand saturation \\
\midrule
Shanghai Pudong & No dispatch & 3445.4 & 39.74 & 36.73 & 6.807 \\
Shanghai Pudong & MPC & 3357.7 & 38.10 & 37.96 & 8.233 \\
Shanghai Pudong & QUBO-inspired SA & 2476.5 & 25.71 & 31.14 & 7.418 \\
Hangzhou Xiaoshan & No dispatch & 2053.3 & 51.71 & 54.85 & 0.936 \\
Hangzhou Xiaoshan & MPC & 2054.6 & 41.16 & 43.20 & 0.976 \\
Hangzhou Xiaoshan & QUBO-inspired SA & 1481.9 & 28.26 & 36.53 & 0.920 \\
\bottomrule
\end{tabular}%
}
\end{table*}

\subsection{Robustness Under Perturbations}
\begin{figure}[h]
\centering
\includegraphics[width=\columnwidth]{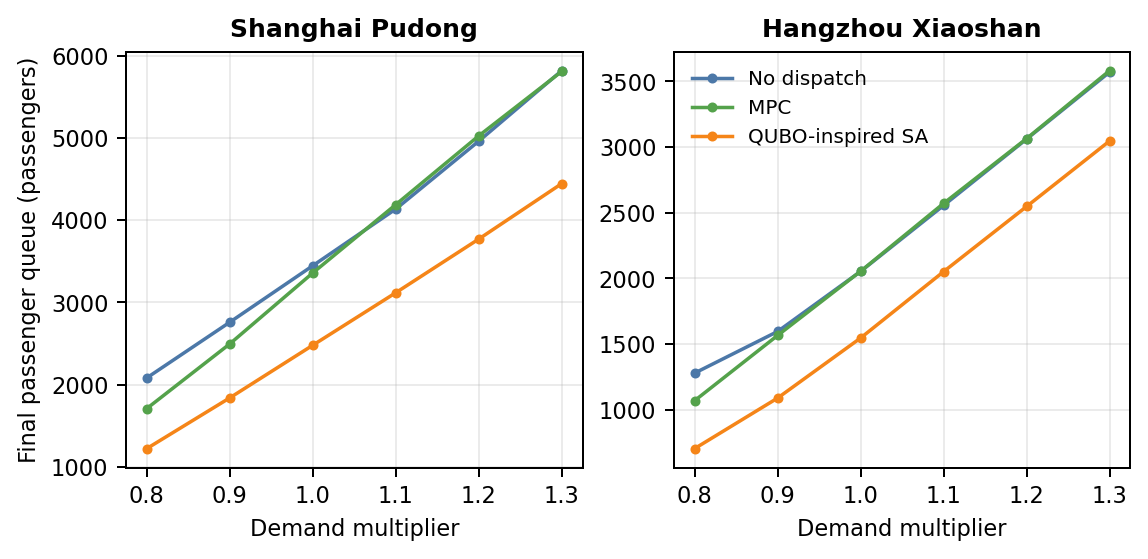}
\caption{Final passenger queue response to demand multipliers.}
\label{fig:robustness}
\end{figure}

Demand-scaling experiments show that final passenger queues increase with demand in all methods. QUBO-inspired simulated annealing retains lower final queues than the other two methods over the tested demand range. When the demand multiplier reaches 1.3, the final passenger queue is 4444.9 passengers at Shanghai Pudong and 3046.3 passengers at Hangzhou Xiaoshan for the QUBO-inspired method. Corresponding no-dispatch values are 5819.0 and 3570.3 passengers. Figure~\ref{fig:robustness} reports the final-queue response to the tested demand multipliers.

\begin{figure}[h]
\centering
\includegraphics[width=\columnwidth]{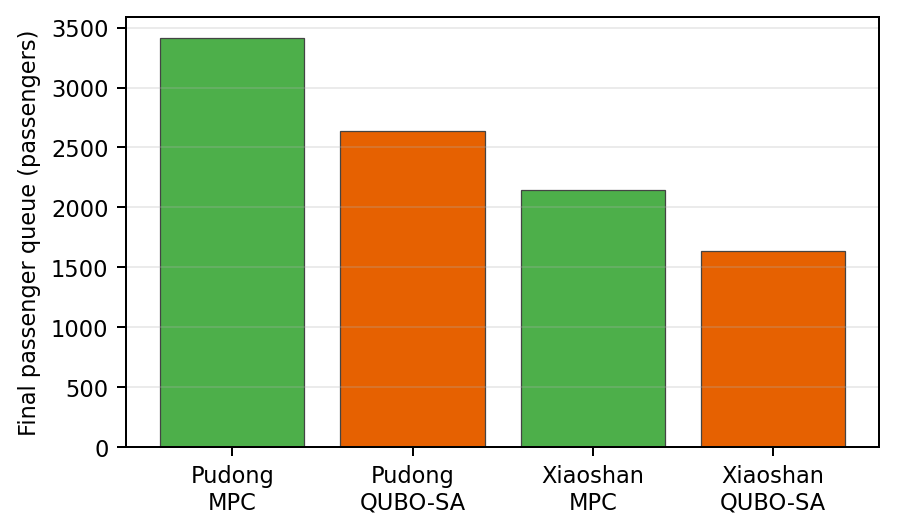}
\caption{Monte Carlo mean final passenger queues at demand noise CV = 0.3.}
\label{fig:mc}
\end{figure}

Random perturbations are also applied to passenger and vehicle arrivals. At a demand noise coefficient of variation of 0.3, the QUBO-inspired method has a final-queue coefficient of variation of 0.119 at Shanghai Pudong and 0.110 at Hangzhou Xiaoshan. The mean final queue remains below the MPC value at both airports. The result does not establish general stability, but it suggests limited sensitivity to moderate random input deviations in the tested scenarios. Figure~\ref{fig:mc} reports the Monte Carlo mean final queues for this noise level.

\subsection{Operational Interpretation}
Dispatch benefits depend on the dominant bottleneck. For Shanghai Pudong, road saturation is the main pressure. Queue clearance may therefore need to be balanced with road protection. A dispatch policy that increases service without considering downstream roadway saturation may move the queue from the terminal to the road. For Hangzhou Xiaoshan, the berth-facility constraint is more influential. Allocating pickup service toward ride-hailing vehicles and improving the passenger-vehicle matching rate can reduce the queue more directly.

QUBO-inspired search evaluates a larger discrete action space and obtains larger queue reductions in the tested cases. MPC has a more explicit interpretation because its candidate actions and horizon costs are transparent. The two methods are therefore treated as complementary. MPC can support transparent operational control, while QUBO-inspired search can be useful when the action space is more combinatorial.

\section{Conclusion and Future Work}
Airport landside congestion is reformulated as an integrated computational problem. A five-minute state model links passenger demand, vehicle supply, pickup berth service, storage capacity, and road capacity. A bottleneck diagnosis layer is then combined with MPC and QUBO-inspired simulated annealing dispatch experiments.

Case results suggest that Shanghai Pudong International Airport and Hangzhou Xiaoshan International Airport can have different dominant landside bottlenecks under the same modeling framework. Shanghai Pudong shows stronger road saturation pressure. Hangzhou Xiaoshan shows stronger pickup berth service pressure, especially for ride-hailing vehicles. Dynamic dispatch reduces final passenger queues and waiting estimates in the tested peak scenarios. QUBO-inspired simulated annealing provides larger queue reductions in the reported experiments, while MPC provides a more structured and interpretable rolling-horizon policy.

Several limitations remain. Some facility parameters are inferred from public information and operational assumptions. More accurate calibration would require field observations of vehicle arrivals, berth occupancy, dwell time, and roadway speed. The BPR-based speed output is treated as a pressure indicator rather than measured traffic speed. Future work can integrate real-time trajectory data, stochastic passenger choice, and airport-specific control rules. Comparison with microscopic simulation and operating records would further improve external validation when such data are available.

\printbibliography

@techreport{ACRP266,
  author = {{National Academies of Sciences, Engineering, and Medicine}},
  title = {Airport Curbside and Terminal Area Roadway Operations},
  institution = {The National Academies Press},
  address = {Washington, DC},
  year = {2024},
  doi = {10.17226/27952},
  url = {https://nap.nationalacademies.org/catalog/27952/airport-curbside-and-terminal-area-roadway-operations}
}

@techreport{ACRP286,
  author = {Le Bris, Gael and Nguyen, Loup-Giang and Atallah, Stephanie and Romero, Karla Medina and Sanchez, Daniel and Tremain, Taylor and Kavaipatti, Prasanna and Nagy, Julia},
  title = {Enhancing Airport Access with Emerging Mobility},
  institution = {The National Academies Press},
  address = {Washington, DC},
  year = {2025},
  doi = {10.17226/28600},
  url = {https://nap.nationalacademies.org/catalog/28600/enhancing-airport-access-with-emerging-mobility}
}

@article{Ugirumurera2021,
  author = {Ugirumurera, Juliette and Severino, Joseph and Ficenec, Karen and Ge, Yanbo and Wang, Qichao and Williams, Lindy and Chae, Junghoon and Lunacek, Monte and Phillips, Caleb},
  title = {A modeling framework for designing and evaluating curbside traffic management policies at Dallas-Fort Worth International Airport},
  journal = {Transportation Research Part A: Policy and Practice},
  volume = {153},
  pages = {130--150},
  year = {2021},
  doi = {10.1016/j.tra.2021.07.013},
  url = {https://doi.org/10.1016/j.tra.2021.07.013}
}

@article{Harris2017,
  author = {Harris, Tyler M. and Nourinejad, Mehdi and Roorda, Matthew J.},
  title = {A Mesoscopic Simulation Model for Airport Curbside Management},
  journal = {Journal of Advanced Transportation},
  volume = {2017},
  pages = {4950425},
  year = {2017},
  doi = {10.1155/2017/4950425},
  url = {https://www.hindawi.com/journals/jat/2017/4950425/}
}

@article{Dong2021,
  author = {Dong, Xiaoxia and Ryerson, Megan S.},
  title = {Taxi Drops Off as Transit Grows amid Ride-Hailing's Impact on Airport Access in New York},
  journal = {Transportation Research Record: Journal of the Transportation Research Board},
  volume = {2675},
  number = {2},
  pages = {74--86},
  year = {2021},
  doi = {10.1177/0361198120963116},
  url = {https://doi.org/10.1177/0361198120963116}
}

@article{Wang2022TransportPolicy,
  author = {Wang, Zi-Jia and Jia, Hui-Hui and Dai, Fangzhou and Diao, Mi},
  title = {Understanding the ground access and airport choice behavior of air passengers using transit payment transaction data},
  journal = {Transport Policy},
  volume = {127},
  pages = {179--190},
  year = {2022},
  doi = {10.1016/j.tranpol.2022.09.001},
  url = {https://doi.org/10.1016/j.tranpol.2022.09.001}
}

@article{Colovic2022,
  author = {Colovic, Aleksandra and Pilone, Salvatore Gabriele and Kukic, Katarina and Kalic, Milica and Dozic, Slavica and Babic, Danica and Ottomanelli, Michele},
  title = {Airport Access Mode Choice: Analysis of Passengers' Behavior in European Countries},
  journal = {Sustainability},
  volume = {14},
  number = {15},
  pages = {9267},
  year = {2022},
  doi = {10.3390/su14159267},
  url = {https://doi.org/10.3390/su14159267}
}

@article{Mahesh2025,
  author = {Mahesh, Srinath and Calvert, Simeon C.},
  title = {Decarbonizing airport access: A review of landside transport sustainability},
  journal = {Transportation Research Part D: Transport and Environment},
  volume = {140},
  pages = {104625},
  year = {2025},
  doi = {10.1016/j.trd.2025.104625},
  url = {https://doi.org/10.1016/j.trd.2025.104625}
}

@article{Liu2023CurbPricing,
  author = {Liu, Jiachao and Ma, Wei and Qian, Sean},
  title = {Optimal curbside pricing for managing ride-hailing pick-ups and drop-offs},
  journal = {Transportation Research Part C: Emerging Technologies},
  volume = {146},
  pages = {103960},
  year = {2023},
  doi = {10.1016/j.trc.2022.103960},
  url = {https://doi.org/10.1016/j.trc.2022.103960}
}

@article{Vishnoi2025,
  author = {Vishnoi, Suyash C. and Simoni, Michele D.},
  title = {Surrogate-based real-time curbside management for ride-hailing and delivery operations},
  journal = {Transportmetrica B: Transport Dynamics},
  volume = {13},
  number = {1},
  year = {2025},
  doi = {10.1080/21680566.2025.2496823},
  url = {https://doi.org/10.1080/21680566.2025.2496823}
}

@misc{CAAC2025,
  author = {{Civil Aviation Administration of China}},
  title = {2025 National Civil Transport Airport Production Statistics Bulletin},
  year = {2026},
  url = {https://www.caac.gov.cn/English/News/202603/t20260304_230166.html},
  note = {Accessed for public airport traffic statistics}
}

@misc{Geofabrik2026,
  author = {{Geofabrik GmbH and OpenStreetMap contributors}},
  title = {OpenStreetMap Data Extracts: China},
  year = {2026},
  url = {https://download.geofabrik.de/asia/china.html},
  note = {Source of Shanghai and Zhejiang road-network extracts}
}

@article{Little1961,
  author = {Little, John D. C.},
  title = {A Proof for the Queuing Formula: {$L = \lambda W$}},
  journal = {Operations Research},
  volume = {9},
  number = {3},
  pages = {383--387},
  year = {1961},
  doi = {10.1287/opre.9.3.383},
  url = {https://doi.org/10.1287/opre.9.3.383}
}

@techreport{BPR1964,
  author = {{Bureau of Public Roads}},
  title = {Traffic Assignment Manual},
  institution = {U.S. Department of Commerce, Urban Planning Division},
  address = {Washington, DC},
  year = {1964},
  url = {https://catalog.hathitrust.org/Record/000968330}
}

@article{Daganzo1994,
  author = {Daganzo, Carlos F.},
  title = {The cell transmission model: A dynamic representation of highway traffic consistent with the hydrodynamic theory},
  journal = {Transportation Research Part B: Methodological},
  volume = {28},
  number = {4},
  pages = {269--287},
  year = {1994},
  doi = {10.1016/0191-2615(94)90002-7},
  url = {https://doi.org/10.1016/0191-2615(94)90002-7}
}

@article{Mayne2000,
  author = {Mayne, D. Q. and Rawlings, J. B. and Rao, C. V. and Scokaert, P. O. M.},
  title = {Constrained model predictive control: Stability and optimality},
  journal = {Automatica},
  volume = {36},
  number = {6},
  pages = {789--814},
  year = {2000},
  doi = {10.1016/S0005-1098(99)00214-9},
  url = {https://doi.org/10.1016/S0005-1098(99)00214-9}
}

@article{Ye2019,
  author = {Ye, Bao-Lin and Wu, Weimin and Ruan, Keyu and Li, Lingxi and Chen, Tehuan and Gao, Huimin and Chen, Yaobin},
  title = {A survey of model predictive control methods for traffic signal control},
  journal = {IEEE/CAA Journal of Automatica Sinica},
  volume = {6},
  number = {3},
  pages = {623--640},
  year = {2019},
  doi = {10.1109/JAS.2019.1911471},
  url = {https://doi.org/10.1109/JAS.2019.1911471}
}

@article{Glover2022,
  author = {Glover, Fred and Kochenberger, Gary and Hennig, Rick and Du, Yu},
  title = {Quantum bridge analytics I: a tutorial on formulating and using QUBO models},
  journal = {Annals of Operations Research},
  volume = {314},
  pages = {141--183},
  year = {2022},
  doi = {10.1007/s10479-022-04634-2},
  url = {https://doi.org/10.1007/s10479-022-04634-2}
}

@article{Lucas2014,
  author = {Lucas, Andrew},
  title = {Ising formulations of many NP problems},
  journal = {Frontiers in Physics},
  volume = {2},
  pages = {5},
  year = {2014},
  doi = {10.3389/fphy.2014.00005},
  url = {https://www.frontiersin.org/articles/10.3389/fphy.2014.00005/full}
}

@article{Kirkpatrick1983,
  author = {Kirkpatrick, S. and Gelatt, C. D. and Vecchi, M. P.},
  title = {Optimization by Simulated Annealing},
  journal = {Science},
  volume = {220},
  number = {4598},
  pages = {671--680},
  year = {1983},
  doi = {10.1126/science.220.4598.671},
  url = {https://doi.org/10.1126/science.220.4598.671}
}

\end{document}